# Unsupervised learning human's activities by overexpressed recognized non-speech sounds

*Serge Smidtas[1], Magalie Peyrot[1]*

[1] Camera-Contact ENoLL Living-Lab, Guéret, France

`{Serge.Smidtas,Magalie.Peyrot}@Camera-Contact.com`

## Abstract

Human activity and environment produces sounds such as, at home, the noise produced by water, cough, or television. These sounds can be used to determine the activity in the environment. The objective is to monitor a person's activity or determine his environment using a single low cost microphone by sound analysis. The purpose is to adapt programs to the activity or environment or detect abnormal situations.

Some patterns of over expressed repeatedly in the sequences of recognized sounds inter and intra environment allow to characterize activities such as the entrance of a person in the house, or a tv program watched.

We first manually annotated 1500 sounds of daily life activity of old persons living at home recognized sounds. Then we inferred an ontology and enriched the database of annotation with a crowed sourced manual annotation of 7500 sounds to help with the annotation of the most frequent sounds.

Using learning sound algorithms, we defined 50 types of the most frequent sounds. We used this set of recognizable sounds as a base to tag sounds and put tags on them.

By using over expressed number of motifs of sequences of the tags, we were able to categorize using only a single low-cost microphone, complex activities of daily life of a persona at home as watching TV, entrance in the apartment of a person, or phone conversation including detecting unknown activities as repeated tasks performed by users.

## 1. Introduction

Programs (software) need to adapt to their environment and the user's activity. Just as geolocalisation became an input of various programs, activity or environment determine by the analysis of the sound, could allow a program to adapt its content to the environment, such as a transportation for example to guide the user by providing knowledge on the railway station or guide the user to its destination, or such as a meeting room to make the system quite.

Programs also require to adapt to the activity of the user such as the entrance in the house of a person as for example by welcoming the person, or command some home automation devices, or such as the reaction of a user to an alarm as for example forwarding an alarm to the user's email or by repeating the alarm if no reaction of the human could be observed letting think that the user did not get the information in time. Other use cases have been proposed based on recognition of the context [1-4] or even monitoring food intake by sound analysis [5] but usually the sound analysis algorithms are combined with other more intrusive sensors for analysis in home environments [6].

At last, the activity of the user is interesting to monitor the user's activity and detect abnormal situation [7] such as a person falling, an intrusion not expected, or even long term variation of habits to detect at an early stage, state of activities that could induced psychological disorders such as boring repeated or at the contrary chaotic activities among weeks by a user or an activity consumed too much as television; or control and supervised worker's activity such as if a house keeping person has correctly perform her job at the right time, duration, by verifying the noise generated such as the one of a vacuum cleaner.

As they are a very large or infinite number of activities and noise produced by these activities, designing models to recognize each activity in each environment is a very long task.

Therefore, we wanted to detect automatically, annotate and recognize unsupervisingly user's activities, as for example in a near context [8,9]. Activities could be detected by looking for recurring sequences of individual noise events. For example, the quasi-absence of noise followed by a hand-door noise, and locker, a creaking door, a door clapping, the noise of the keychain and lock should infer the activity that a person is interring his house. As these noises individually or in a miss-order won't have the same meaning. This activity, defined by this set of sequenced single noises, repeated quasi every day will allow to notice and define it without the requirement of any configuration of the system but by requiring an initial duration of observation by the system.

Other examples in a daily living house activity in the context of old persons are: phoning, that should be revealed by the repetition of events in which there are alternative moment of speech and silence (when the user is listening to his contact), and with a probability increased if a phone ring was heard before the conversation started.

Most studies concerning recognition of human activities at home require a good knowledge on the activities that on want to recognize [10-12], including using only sound analysis [13-15]. For a survey on general machine recognition of human activity see [16].

We applied this objective in the context of older persons +75 living alone at home during months. This offers a stable environment homogeneous, in the framework of a French national research project SweetHome[17]. Related work has also been done in the home environment.

## 2. Method

To record sounds in home environment, we used the VisAge platform (http://camera-contact.com) designed specifically for elders and isolated persons. The platform is composed of a connected touch screen. Sounds in the environment of the house of 14 persons for about 5 months were studied.

First we needed to know the variety of single sounds events that compose the daily life of our elders.

For this purpose, we first start with a set of recognizable sound for which previous models have been found. These sounds models have been provided by Sehili et al.[18]. We designed interactive programs to encourage VisAge volunteer users to record these sounds in their home environment (figure 1).

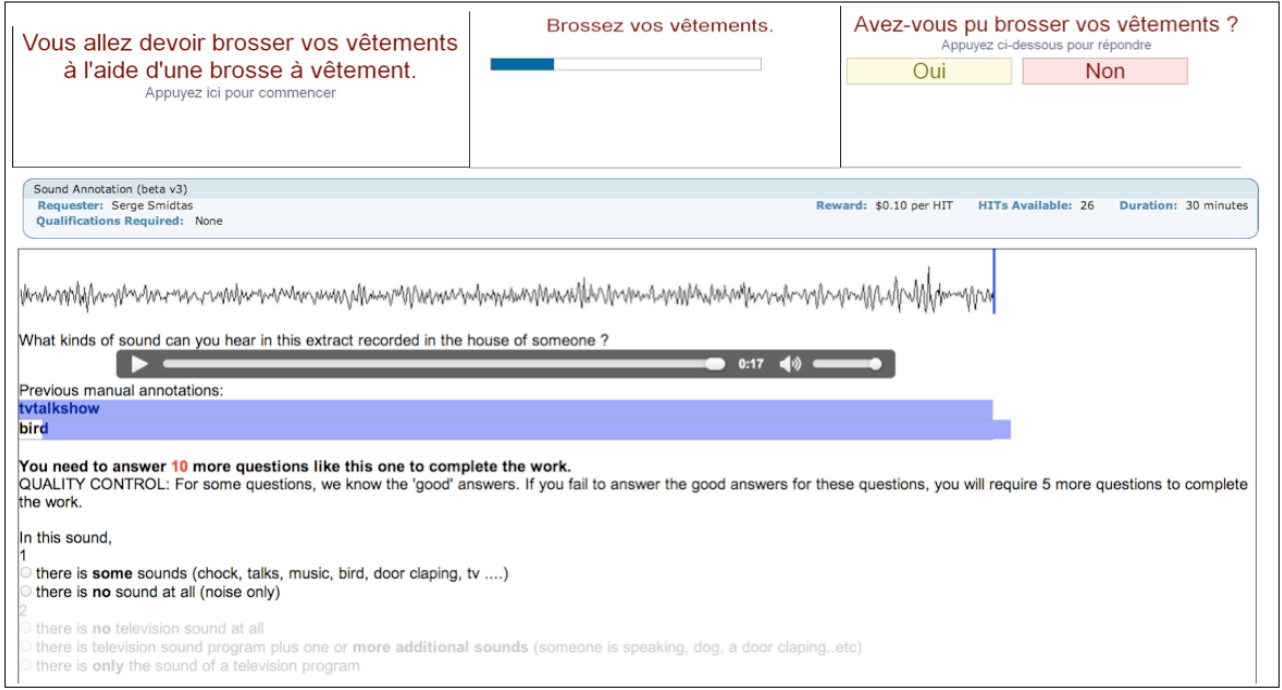

Figure 1 *Top: Interactive screens broadcasted on VisAge screen asking persons to volunteering produce sounds. "1-You will be asked to brush you clothes. 2-Brush you clothes. 3-Have you brushed your clothes, yes, No". Bottom: Annotating tool for using the Amazon Mturk crowed-sourcing tool.*

Unfortunately, we have been confronted to the limit of this method: the sounds were not the most frequent ones in the environment of our users, and most of the type of sounds recognizable did not make sense at all for our users to produce them.

To focus on the most frequent sounds we recorded the sounds in the real environment using the VisAge living-lab. To eliminate the period where there were no sound at all, we developed an ad-hoc algorithm to record sound on variation of sound levels detection. More than 100k sounds with an average duration of 30 seconds were used in 11 environments, representing more than a month of interesting sounds to analyze.

The environment was composed of the living room (either the kitchen or the living room) of 11 houses of persons living mostly alone in France and mostly old (median is 85 years old).

Crowd-sourcing using Amazon Mechanical Turk (mturk.com) was used to annotate the database. A previously based of 1700 annotations were used for quality check of the crowd source results. 30 annotators among the word amplified this by adding qualifying 17.000 annotations (figure 1). These annotations included some descriptive knowledge, splitting and categorizing the sounds in classes of the Visage Sound Ontology (http://camera-contact.com/vso.rtf, see figure 2).

Then the set of annotated sounds have been used to measure and characterize the diversity of the interesting sounds and 50 tags (models), also called models, of the sounds have been established using the learning algorithms of Sehli et al., but without considering their semantic knowledge to cover the space (for instance, the sounds are tagged as sound A and sound B instead of sound of a dog barking and sound of a bird song).

Activities are composed of a chain of short audible events and the pattern of these tags can be found and characteristic of the activity. Theses chains can be detected to be repeat themselves in an environment, within a home or among homes. For example, in all home environments, the entrance of a person in a house is composed by the noise of the door knock, followed by the door clapping. Some event may occur also in only a single environment, for example, you may not decorate your living room every day then the sound of drill is not a daily sound.

To look for such chains of sound events characterizing an activity, we looked for overexpressed complex motifs [19].

a) We looked for period of time dt during which several couple of tags $(f_i, f_j) \in F_i \times F_j$ occur more frequently than expected.

$$\forall\ i,j \in \{1,50\}^2,\ p((f_i{}^t_{t+dt}, f_j{}^t_{t+dt})) \propto O(Card_i(F_i) \cdot Card_j(F_j))$$

b) We looked for tags of sounds that follow each other with a probability above than expected after a delay dt and also for the one for which the following is ordered more frequently in one order than the other as they are expected to be the same.

$$p(f_i{}^t_{t+dt} | f_j{}^{t+\partial}_{t+\partial+dt}) \propto O(Card_i(F_i) \cdot Card_j(F_j))|_{\partial=dt=cst}$$
$$p(f_i{}^t_{t+dt} | f_j{}^{t+\partial}_{t+\partial+dt}) > p(f_j{}^t_{t+dt} | f_i{}^{t+\partial}_{t+\partial+dt})$$

The objective of the blast/scalar product of the sequences, is to put together events such as watching TV, entering the house, making cleaning, doing human activities, or eating.

## 3. Results

We built a hierarchical ontology of every-day sound classes based on the most frequent sound of elders' home environment. It has been used to compare activities inferred with activities making sense for us (figure 2).

- Thing
  - chocrecognized
    - dishes
    - drawer
      - zip
    - frictiondrawn
    - metallic
      - glassbreaking
      - keys
      - lock
    - paper
    - wood
      - doorclaping
        - doorgrinding
  - corruptedfile
  - notcorrupted
    - noiseonly
    - nosound
    - sounds
      - animals
        - bird
        - dog
          - dogbarking
          - dogcomplaint
      - artefact
      - humanactivity
        - interactionwithobjects
          - activity
            - housecleaning
          - clickbutton
          - keyboard
          - tissue
            - clothingshake
            - zip
        - movement
          - walking
        - speech
          - childspeach
          - discussion
            - brawlvocal
            - phoning
          - emotion
            - Angry
            - Cry

Figure 2 *Extract of the ontology*

The ontology is provided here at http://camera-contact.com/CameraContactSoundOntology.rdf .

We built a crowd sourced annotated tool, source code available on request by academics (https://www.mturk.com/mturk/preview?groupId=255CF0CP5KXPZL4Z1HPG8PM7NK9WXU ) to annotate at large scale the sound and activities.

We used then 300k sounds (including the first 100k set) to learn the re-occuring events, and we used the 17k manually annotated sounds as a quality control and wording of the events.

We found several activities based on repeated tags and tags that sequentially not randomly appear.

We used the following values: dt=∂=1 minute, and we tag the sounds with 319.204 instances of tags.

To illustrate the results, following are some examples for which knowledge of the original reason of choosing a tag could make sense.

For example, the tag DogComplaints : The name of this tag do not mean that the tag characterize dog complaints, but only that the tag has been first found in dog complaints sounds.

10% more frequently the tag labeled DogComplaints occurs some seconds right after the tag WomenSpeach as if we could infer that dog complaints is a respond from dog to the speech of someone as described in communication between wolves and human [20]. The tag DogComplaints followed by a WomenSpeach was recorded 8581 times (vs 9980 for the opposite order) during the months of the experiment.

A second example is the tag GlassBreaking that occurs followed by the tag WomanSpeech 50% more frequently in that order for more than 1473 instances. As one can expect, after we have broken something, this make someone react and say usually bad words.

The TV tag occurrences stop 30% more frequently just after a

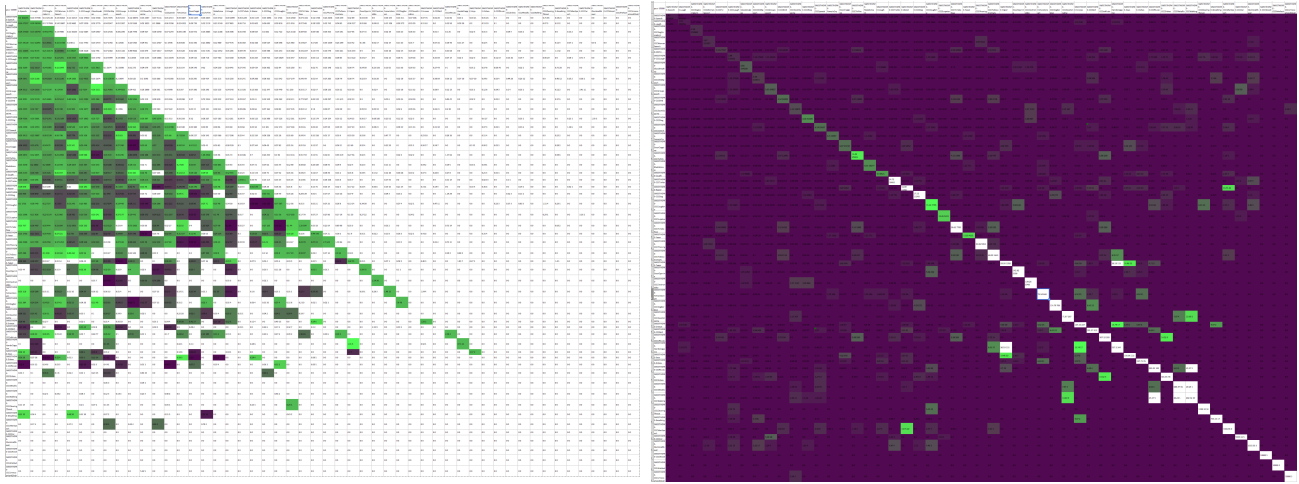

Figure 3 *Each column or line represents one of the 50 tags. a) Left, is the matrix of the frequency of tags found at the same time dt=1min. This matrix indicates the frequency of each tag, its repetition frequency within the diagonal terms, and in the triangular terms the co-occurring of the tags. The matrix is symmetric. Green terms are the values that distinct from the expected values, which means that sounds are often repeated.*

*b) Right, is the matrix of the frequency of the tags following one another after a ∂ delay of 1 minute. In columns, these are the first tags found. In rows are the tags following the first tag. The matrix is not symmetric. Green terms are the values close to their symmetric term. Light green and purple are the terms that distinct the most from their symmetric term.*

Ring tag than just before, as we can imagine that someone answering the phone turn off the TV to speak.

On figure 3-a, other results of repeated tags are shown and 3-b sequentially tags. These matrixes can be seen as directed graphs that allow to search graph patterns that occur more or less frequently than random to find knowledge. The 3 examples given above represent each, simple edges, but loop or even more complex topological motifs can be search (as it has been done in graphs in other similar domain [19].

Application: We used the pattern of some events to adapt the broadcasted program on the touch screen terminal to this context. Two examples of services: When TV is on and its sound can be heard, we automatically broadcast the augmented-TV program on the VisAge screens. When someone enters his house, we broadcast a welcoming message on the screen. These are some examples of programs adapted without filling any intrusion sentiment for the users.

## 4. Discussion

More various graph pattern and chains could be searched. Here we only gave a sketch and the general ideas. In particular, AI and pattern matching techniques, clustering, biclustering, Markov chains could be used to detect event made by several sounds that do not appear sequentially at random. Then these events could be labeled and knowledge is produced. We have sketch a first draft of a methodology to explore the grammar of the non-word sounds of every day life that could be used to make prediction on behavior [21]. These words combined in certain logic make sentences, and here we could say activities.

Listening to a ring, then stopping the radio, and talking on phone make sense as one activity action of answering phone for example. The need to recognize and monitor activities of elders at home is very important to adapt tools and complement their lack of capacity and motivation. More specific research should be based on unsupervised recognition of activities and rely on the social and cloud effect to determine by comparison what an *activity* is, and that could be defined as a pattern re-occurring to a person or among persons/homes. These algorithms should rely on sensors and devices that are not intrusive also for the house, such as a simple microphone embed in a box, and unlike IR sensors that required to be installed in numbers in the houses and that are for this reason less acceptable for the end-users and deterioration of their old houses.

## 5. Acknowledgements

We would like to thank Mohamed Sehili and Dan Istrate for their tagging tools. This work has been cofounded by the French National Research Agency (ANR-09-VERS-011).

## 6. References

[1] Clarkson, B., & Pentland, A. "Extracting context from environmental audio". In Wearable Computers, 1998. Digest of Papers. Second International Symposium on (pp. 154-155). IEEE 1998

[2] Eronen, A. J., Vesa T. Peltonen, Juha T. Tuomi, Anssi P. Klapuri, Fagerlund S., Sorsa T., Lorho G., Huopaniemi J.. "Audio-based context recognition." Audio, Speech, and Language Processing, IEEE Transactions on 14, no. 1 (2006): 321-329.

[3] Heittola, T, Mesaros A., Eronen A., Virtanen T. "Audio context recognition using audio event histograms." In Proc. of the 18th European Signal Processing Conference (EUSIPCO 2010), pp. 1272-1276. 2010

[4] Peltonen V. , Tuomi J., Klapuri A, Huopaniemi J., Sorsa T.. "Computational auditory scene recognition." In Acoustics, Speech, and Signal Processing (ICASSP), 2002 IEEE International Conference on, vol. 2, pp. II-1941. IEEE, 2002.

[5] Passler, S., and Fischer. W-J. "Food Intake Activity Detection Using a Wearable Microphone System." In Intelligent Environments (IE), 2011, pp. 298-301. IEEE, 2011.

[6] Brdiczka, O., Langet M., Maisonnasse J., Crowley J.L. "Detecting human behavior models from multimodal observation in a smart home." Automation Science and Engineering, IEEE Transactions on 6, no. 4 (2009): 588-597.

[7] Jakkula, V., and Cook. D.J. "Detecting Anomalous Sensor Events in Smart Home Data for Enhancing the Living Experience." Artificial Intelligence and Smarter Living 11 (2011): 07.

[8] Flanagan J. A., Mäntyjarvi J., Himberg J.. 2002. "Unsupervised Clustering of Symbol Strings and Context Recognition". 2002 IEEE (ICDM '02), Washington, DC, USA, 171.

[9] Park, A. S., and Glass J.R. "Unsupervised pattern discovery in speech." Audio, Speech, and Language Processing, IEEE Transactions on 16, no. 1 (2008): 186-197.

[10] Sigg S., Scholz M., Shi, S., Ji Y., Beigl M., "RF-Sensing of Activities From Non-Cooperative Subjects in Device-Free Recognition Systems Using Ambient and Local Signals" IEEE Transactions on Mobile Computing, 2013. IEEE.

[11] Chahuara P., Fleury A., Portet F., Vacher M., "Using Markov Logic Network for On-Line Activity Recognition from Non-visual Home Automation Sensors", Ambient Intelligence LNCS 7683 (2012) 177-192

[12] Nuria O., Horvitz E., Garg A. "Layered representations for human activity recognition." In Multimodal Interfaces, 2002. Proceedings. Fourth IEEE International Conference on, pp. 3-8. IEEE, 2002.

[13] Stork, J. A., Spinello L., Silva J., and Arras K.O.. "Audio-based human activity recognition using Non-Markovian Ensemble Voting." In RO-MAN, 2012 IEEE, pp. 509-514. IEEE, 2012.

[14] Mesaros A., Heittola T., Eronen A., Virtanen T.. "Acoustic event detection in real life recordings." In 18th European Signal Processing Conference, pp. 1267-1271. 2010.

[15] Zheng, H., Wang H., Black N.. "Human activity detection in smart home environment with self-adaptive neural networks." In Networking, Sensing and Control, 2008. ICNSC 2008. IEEE International Conference on, pp. 1505-1510. IEEE, 2008.

[16] Turaga, P., Chellappa R., Subrahmanian V. S., and Udrea O.. "Machine recognition of human activities: A survey." Circuits and Systems for Video Technology, IEEE 18, no. 11 (2008): 1473-1488

[17] Vacher, M., Istrate M., Portet F., Joubert T., Chevalier T., Smidtas S., Meillon B. Lecouteux B., Sehili M., Chauara P., Méniard S. "The sweet-home project: Audio technology in smart homes to improve well-being and reliance." In Engineering in Medicine and Biology Society, EMBC, 2011, pp. 5291-5294. IEEE.

[18] Sehili, M. A., Lecouteux, B., Vacher, M., Portet, F., Istrate, D., Dorizzi, B., & Boudy, J. (2012). "Sound Environment Analysis in Smart Home. In Ambient Intelligence" (pp. 208-223). Springer Berlin Heidelberg.

[19] Smidtas S, Yartseva A., Schächter V., Képès F.. "Model of interactions in biology and application to heterogeneous network in yeast." CR Biologies 329, no. 12 (2006): 945-952.

[20] Hartmann Jenal/Hildegard Hoppe: Wolfsfreund - Werner Freund und seine Wölfe. Bildband. Conte Verlag 2006..

[21] Lee S.W., Kim Y. S., Bien Z., "Learning Human Behavior Patterns for Proactive Service System: Agglomerative Fuzzy Clustering-Based Fuzzy-State Q-learning," CIMCA, pp.362-367, 2008